\documentclass[10pt]{article} 
\usepackage[preprint]{tmlr}

\usepackage{multirow}
\usepackage{amsmath}

\usepackage{booktabs}    
\usepackage{graphicx}    
\usepackage{subcaption}  
\usepackage[table,xcdraw]{xcolor}  

\usepackage{amsmath,amsfonts,bm}

\def\eqref#1{equation~\ref{#1}}

\def\1{\bm{1}}

\DeclareMathAlphabet{\mathsfit}{\encodingdefault}{\sfdefault}{m}{sl}
\SetMathAlphabet{\mathsfit}{bold}{\encodingdefault}{\sfdefault}{bx}{n}

\usepackage{hyperref}
\usepackage{url}

\usepackage[ruled,vlined]{algorithm2e}

\title{UniGADD: Universal Generator-Agnostic Deepfake Detector}

\author{\name Bernard Tomczyk \email b.tomczyk@lancaster.ac.uk \\
      \addr School of Computing and Communications, Lancaster University\\
      Lancaster, United Kingdom
      \AND
      \name Agil Aghasanli \email a.aghasanli1@lancaster.ac.uk \\
      \addr School of Computing and Communications, Lancaster University\\
      Lancaster, United Kingdom
      \AND
      \name Plamen Angelov \email p.angelov@lancaster.ac.uk \\
      \addr School of Computing and Communications, Lancaster University\\
      Lancaster, United Kingdom}

\begin{document}

\maketitle

\begin{abstract}
In this paper, we introduce Universal Generator-Agnostic Deepfake Detector (UniGADD), a novel method that employs supervised learning to achieve high accuracy on known deepfake generators, while maintaining robust performance on previously unseen ones. The proposed approach follows a two-stage optimisation process. In the first stage, a contrastive loss encourages the model to learn discriminative feature embeddings from real and known fake images, resulting in strong performance within the training domain. In the second stage, the embedding space is refined by promoting inter-cluster separation and intra-cluster compactness, applied exclusively to real samples. This refinement enhances generalisability, enabling the method to exhibit improved robustness against unseen deepfake generation techniques. UniGADD achieves accuracy on par with state-of-the-art methods for known generators, while significantly outperforming them on unseen cases, demonstrating its scalability and practical applicability for adversarial content detection.
\end{abstract}

\noindent\textbf{Code:} \url{https://github.com/Aghasanli/UniGADD-Universal-Generator-Agnostic-Deepfake-Detector}

\section{Introduction}

The rapid proliferation of deepfake generators poses an escalating challenge for forensic tools \citep{deepfakeSurvey2}. Most published detectors are trained with labelled examples from a fixed set of generators, achieving high accuracy on those seen attacks yet faltering when confronted with novel manipulation pipelines. Pellicer \textit{et al.}\ \citep{Pellicer_2024_CVPR} frame the task as unsupervised anomaly detection, showing that generator-agnostic cues exist, although their reliance on nominal-only statistics leaves recent advances in representation learning under-explored.

To address the cold-start scenario where no fake examples are available, we refine foundation-model embeddings with a dedicated contrastive loss and derive prototypes that map the real-image distribution. Samples that deviate from these prototypes are flagged as potential anomalies, enabling detection without generator-specific supervision. This approach sidesteps the need for curated fake datasets and remains robust as synthesis methods evolve.

While such unsupervised strategies excel in generalisability, purely nominal training can sacrifice closed-set precision. Conversely, supervised detectors capture fine-grained artefacts but overfit to the generators present in the training set \citep{DeepfakeSurvey}. We propose the \textbf{Universal Generator-Agnostic Deepfake Detector (UniGADD)}, which unifies these complementary strengths. UniGADD first exploits labelled fakes, when available, to carve a discriminative margin between real and manipulated imagery, then tightens the real-image region through the prototype-based refinement introduced above. The result is a detector that preserves high precision on known attacks while degrading gracefully on unseen ones.

Two practical considerations guide our design. Authentic imagery is plentiful, yet collecting labelled samples for every new generator is unrealistic \citep{DeepfakeAttacks}. At the same time, detectors must remain reliable as synthesis technology evolves. By combining light supervision with feature-space regularisation, UniGADD meets both requirements without continual retraining.

We benchmark UniGADD on the AIGC datasets \citep{fan2024aigcbench}, which contain a diverse mixture of GAN- and diffusion-based generators and reflect varied real-world usage scenarios \citep{zhu2023genimage}. UniGADD matches or surpasses specialist baselines on their home domains and maintains competitive performance when confronted with previously unseen generators, supporting its suitability for real-world deployment.

Our main contributions are as follows:
\begin{itemize}
    \item We introduce UniGADD, a contrastive clustering framework that generalizes synthetic image detection to unseen generative models.
    \item We propose a real-image clustering strategy that enables the model to leverage intrinsic data structure without supervision.
    \item We establish a leave-one-out protocol covering both GAN- and diffusion-based generators, enabling rigorous evaluation of cross-domain transfer.
    \item We demonstrate competitive performance on the challenging AIGC benchmark, highlighting strong one-to-many generalization.
\end{itemize}

\paragraph{Terminology.} 
Throughout this paper we use the term ``deepfake'' in a broad sense to mean AI-generated visual content (AIGC). We adopt this wording because ``deepfake'' has become the prevalent public term for generated images, but whenever precise distinctions are required we explicitly refer to GAN- or diffusion-based synthesis.

\section{Related Work}
Early forensic efforts relied on handcrafted cues such as lighting, facial landmarks, or head pose \citep{farid2016photo,sencar2013digital}.  As GANs and diffusion models advanced, supervised classifiers trained on curated real/fake datasets became standard \citep{faceforensics,yang2019exposing,ijcai2020p476,bird2023cifakeimageclassificationexplainable}, with DE-FAKE \citep{sha2022fake} and the contrastive method of Baraldi \textit{et al.}\ \citep{baraldi2024contrasting} demonstrating strong closed-set accuracy.  These systems, however, tend to generalise poorly to unseen generators, as documented by Ojha \textit{et al.}\ \citep{ojha2023towards}.  Frequency-domain studies report similar limitations, with artefacts learnt from GAN families rarely transferring to diffusion models \citep{frank2020leveraging,wang2020cnn}.

Generator-agnostic detection aims to mitigate this brittleness. Epstein \textit{et al.} \citep{epstein2023online} show substantial accuracy drops when a novel generator appears in a continual-learning setting, and Corvi \textit{et al.} \citep{Corvi_2023_CVPR,corvi2023detection} confirm that detectors trained on GAN artefacts do not automatically extend to diffusion artefacts. Prototype-based detectors, exemplified by PUDD \citep{Pellicer_2024_CVPR}, measure similarity to learned class prototypes and flag low-score inputs as potential fakes, improving robustness to unseen attacks. Recent work couples contrastive objectives with large vision encoders to capture both discrimination and generalisation: Larue \textit{et al.} \citep{larue2023seeablesoftdiscrepanciesbounded} compact the real cluster with a bounded loss, Sun \textit{et al.} \citep{sun2023contrastivepseudolearningopenworld} use pseudo-labels for open-world attribution, and Shi \textit{et al.} \citep{shi2023realfacefoundationrepresentation} report that linear probes on CLIP features can outperform many bespoke detectors, although fine-tuning still offers gains.

Visual anomaly-detection methods model the nominal data distribution and treat deviations as outliers. PatchCore \citep{roth2022totalrecallindustrialanomaly} stores patch features in a memory bank, and PaDiM \citep{defard2020padimpatchdistributionmodeling} employs pixel-wise Gaussian models.  While effective in industrial settings, these designs entail memory overhead or pixel-level supervision that does not transfer cleanly to deepfake tasks. Foundation-model embeddings from CLIP \citep{radford2021learningtransferablevisualmodels} and DINOv2 \citep{oquab2024dinov2learningrobustvisual} provide a strong baseline, yet adapting them without sacrificing general utility remains challenging; Han \textit{et al.} \citep{han2024generalvideobaseddeepfakedetection} show that fine-tuned variants can overfit the target domain.

Prototypical networks compute class centroids in the embedding space and classify samples by distance \citep{Snell2017PrototypicalNF}. Their success in few-shot learning and interpretability has motivated extensions to meta-learning and self-supervised representation learning \citep{Chen2018ThisLL,ANGELOV2025129464,Donnelly2021DeformablePA,Wu2021ProtoTransformerAM}. SwAV \citep{Caron2020UnsupervisedLO} and DINO \citep{Caron2021EmergingPI} organise unlabelled data into semantically meaningful clusters, although prototype collapse can limit diversity, prompting regularisation strategies such as ReSA \citep{Weng2025ClusteringPO}. The contrastive loss used in UniGADD builds on this body of work by enforcing inter-prototype separation and intra-prototype compactness within the real-image manifold, enabling efficient anomaly cues while retaining the general features of the underlying foundation model.

UniGADD integrates supervised contrastive learning, prototype-based refinement, and lightweight anomaly detection, preserving high precision on generators seen during training and improves recall on unseen ones without requiring extensive memory resources or continual retraining.

\section{Proposed Method}

\begin{figure*}[h!]
    \centering
    \includegraphics[width=0.8\textwidth]{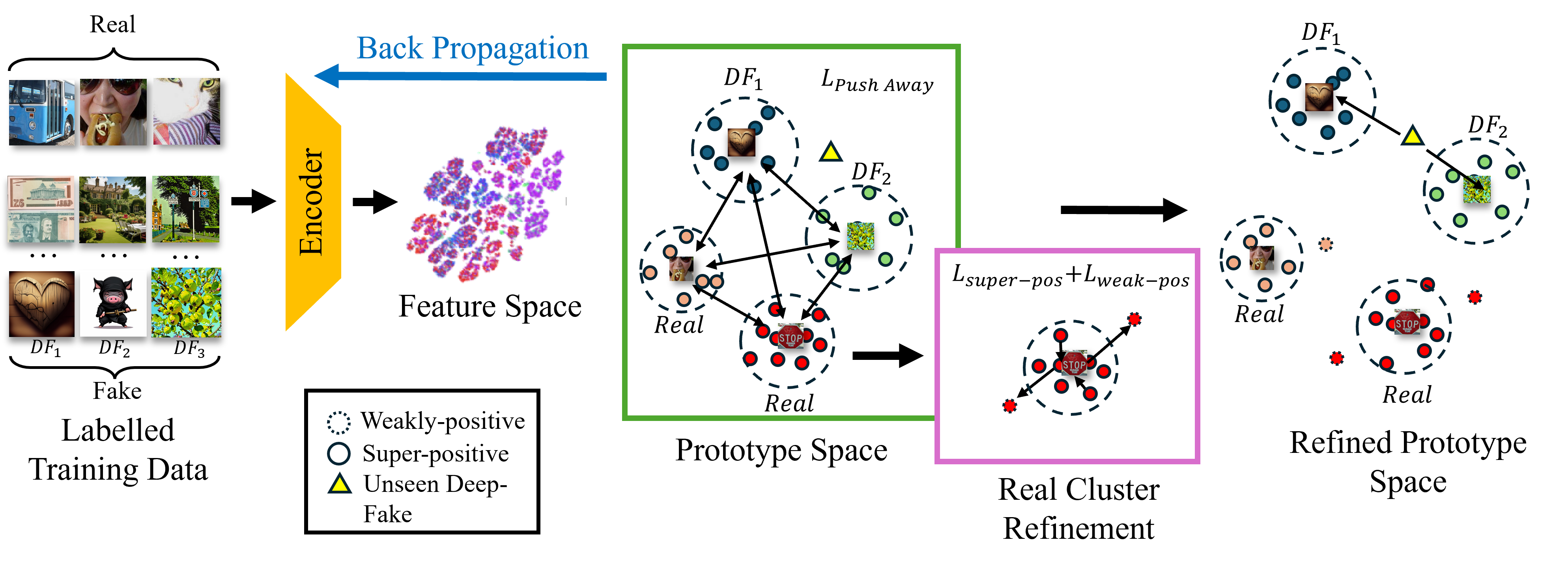}
    \caption{Overview of the proposed method, consisting of feature extraction, supervised contrastive prototype refinement and anomaly detection.}
    \label{fig:method}
\end{figure*}

We propose a supervised contrastive learning framework for deepfake detection and generator attribution that combines structured pseudo-labelling of real data with explicit supervision from known fake generators. UniGADD follows a two-stage process: the first learns a discriminative embedding space through hybrid labelling, while the second refines cluster structure within the real image distribution to improve robustness and generalisability.

\subsection{Feature Extraction and Prototype Initialisation}\label{test}

Given a set of real and fake training samples \( X = \{ \mathbf{x}_i \}_{i=1}^{N} \), we extract feature embeddings using state-of-the-art Vision Transformer architectures, including DINOv2~\citep{oquab2024dinov2learningrobustvisual} and CLIP~\citep{clip}. DINOv2 produces 1024-dimensional embeddings, while CLIP generates 768-dimensional representations. These embeddings define a unified representation space for both real and fake data, providing a solid foundation for contrastive learning and structural analysis.

We begin by applying K-means clustering over the real sample embeddings to induce an initial structure in the feature space. These cluster assignments serve as pseudo-labels for real samples, while fake samples retain their known generator labels. The combined labels are used to supervise contrastive learning, which sharpens the representation structure by encouraging alignment within clusters and separation between them. Crucially, this sharpening lays the foundation for the second stage, where we explicitly refine the real clusters through intra-cluster compactness objectives. After contrastive training, we perform prototype selection by identifying the samples closest to each cluster centre in the refined embedding space. This deferred selection ensures that the prototypes reflect the improved geometry of the learned space, leading to enhanced interpretability and robustness in downstream detection and attribution.

\subsection{Stage 1: Supervised Contrastive Learning}\label{subsec:supcon}

To learn a discriminative embedding space, we optimise a supervised contrastive loss~\citep{khosla2020supervisedcontrastivelearning} over a composite label set. Let \( \mathbf{z}_i \) denote the normalised embedding of sample \( i \), and let \( y_i \) be its composite label. The loss is defined as:
\begin{equation}
    L_{\text{contrast}} = \sum_{i=1}^{N} \frac{-1}{|\mathcal{P}(i)|} \sum_{p \in \mathcal{P}(i)} \log \frac{\exp(\mathbf{z}_i \cdot \mathbf{z}_p / \tau)}{\sum_{a \in \mathcal{A}(i)} \exp(\mathbf{z}_i \cdot \mathbf{z}_a / \tau)},
\end{equation}
where \( \mathcal{P}(i) = \{j \neq i \, |\, y_j = y_i \} \) denotes the set of positives for anchor \( i \), \( \mathcal{A}(i) = \{j \neq i \} \) is the full batch excluding the anchor, and \( \tau \) is a temperature parameter.

This objective encourages embeddings of samples with the same composite label, either real samples sharing a pseudo-cluster label or fake samples from the same generator, to be drawn together, while pushing apart dissimilar instances. Crucially, the pseudo-labels for real samples, derived from K-means clustering, are periodically updated during training to reflect the evolving geometry of the feature space. This dynamic relabelling reinforces the underlying structure and helps sharpen cluster boundaries, setting the stage for more effective intra-cluster refinement.

\subsection{Stage 2: Intra-Cluster Compactness}
\label{subsec:refine}

Following supervised contrastive learning, we further refine the structure of the real sample distribution by explicitly enforcing intra-cluster compactness. The pseudo-clusters induced in Stage 1 provide an initial structural prior over the real data. However, these clusters may contain internal variation, and loosely aligned samples near the periphery can obscure clear decision boundaries, especially when facing previously unseen deepfakes.

To address this, we define a two-part objective that consolidates cluster structure by tightening the core and softly regulating the boundary. For each real cluster \( k \), we compute a prototype vector \( \boldsymbol{\mu}_k \) as the mean embedding of the assigned samples. Cosine similarity is then used to assess alignment between a sample and its cluster centre:
\begin{align}
    s_k(\mathbf{z}_i) &= \frac{\mathbf{z}_i \cdot \boldsymbol{\mu}_k}{\|\mathbf{z}_i\| \|\boldsymbol{\mu}_k\|}, \\
    X_k^{(sp)} &= \{ \mathbf{z}_i : y_i = k, \; s_k(\mathbf{z}_i) \geq \mu_k - 2\sigma_k \}, \\
    X_k^{(wp)} &= \{ \mathbf{z}_i : y_i = k, \; s_k(\mathbf{z}_i) < \mu_k - 2\sigma_k \},
\end{align}
where \( \mu_k \) and \( \sigma_k \) denote the mean and standard deviation of similarity scores within cluster \( k \).

The super-positive region \( X_k^{(sp)} \) represents highly aligned, clean real samples. By pulling them closer to the prototype, we reinforce cluster cohesion:
\begin{equation}
    L_{\text{super-pos}} = \sum_{k=1}^{K_r} \sum_{\mathbf{z}_i \in X_k^{(sp)}} \left( 1 - s_k(\mathbf{z}_i) \right)^2.
\end{equation}
Conversely, the weak-positive region \( X_k^{(wp)} \) captures samples that deviate from the cluster core. These peripheral areas are likely to overlap with novel or unseen fake samples in the test phase. To improve anomaly sensitivity, we gently push such samples outward using:
\begin{equation}
    L_{\text{weak-pos}} = \sum_{k=1}^{K_r} \sum_{\mathbf{z}_i \in X_k^{(wp)}} s_k(\mathbf{z}_i).
\end{equation}

The total loss is given by:
\begin{equation}
    L_{\text{refine}} = \lambda_1 L_{\text{super-pos}} + \lambda_2 L_{\text{weak-pos}},
\end{equation}
where \( \lambda_1 \) and \( \lambda_2 \) control the relative contribution of each term.

This refinement stage not only improves the clarity and robustness of real clusters but also increases the likelihood that unseen fake samples will fall outside the compacted real distribution, enabling more accurate detection and attribution.

\subsection{Deepfake Detection and Attribution}

Following the two-stage training process, we compute final cluster centroids for both real and fake classes. These centroids summarise the distributional structure of the training data in the embedding space, where real clusters have been compacted through refinement, and fake clusters are formed based on known generator groupings established during contrastive learning.

For a test embedding \( \mathbf{z}_{\text{test}} \), we calculate its Euclidean distance to each cluster centroid. The sample is classified as real if it is closer to any real cluster than to the closest fake cluster:
\begin{equation}
\text{label} = 
\begin{cases}
\text{Real}, & \text{if } \min_k \|\mathbf{z}_{\text{test}} - \boldsymbol{\mu}_k^{\text{real}}\| < \min_j \|\mathbf{z}_{\text{test}} - \boldsymbol{\mu}_j^{\text{fake}}\|, \\
\text{Fake}_{j^*}, & \text{otherwise, with } j^* = \arg\min_j \|\mathbf{z}_{\text{test}} - \boldsymbol{\mu}_j^{\text{fake}}\|.
\end{cases}
\end{equation}

This nearest-centroid classification mechanism enables simultaneous deepfake detection and generator attribution. The refined structure of real clusters ensures a compact decision region, increasing sensitivity to out-of-distribution manipulations. Meanwhile, fake centroids, which are anchored by generator-specific labels, allow for accurate attribution among known manipulations.

Importantly, since weakly aligned real samples have been softly pushed away from the cluster cores, previously unseen fake content is more likely to fall outside the refined real distribution. This results in improved robustness against generalisation failures and supports open-set detection capabilities in realistic forensic settings.

\begin{algorithm}[h!]
\DontPrintSemicolon
\SetAlgoLined
\SetNoFillComment 
\caption{Supervised Contrastive Detection Framework}
\label{alg:supergadd}

\KwIn{Embeddings, ground-truth generator labels, number of real clusters \( K_r \), learning rate \( \eta \)}

\small{\tcp{Stage 1: Contrastive embedding learning}}
\For{epoch \( = 1 \) to \#Stage 1 epochs} 
{
    Cluster real samples using k-means \;
    Assign composite labels (real pseudo-labels + fake generator labels) \;
    Compute contrastive loss \( L_{\text{contrast}} \) \;
    Update projection model using gradient descent \;
}

\small{\tcp{Stage 2: Cluster refinement on real data}}
\For{epoch \( = 1 \) to \#Stage 2 epochs} 
{
    Compute real cluster prototypes \( \boldsymbol{\mu}_k \) \;
    Identify \( X_k^{(sp)} \), \( X_k^{(wp)} \) via cosine similarity \;
    Compute \( L_{\text{super-pos}} \), \( L_{\text{weak-pos}} \) \;
    Update model using gradient descent \;
}

\small{\tcp{Inference and attribution}}
Store final centroids for real and fake clusters \;
Classify test samples by nearest-centroid comparison \;

\end{algorithm}

\begin{figure*}[h!]
    \centering
    \begin{minipage}[t]{0.32\textwidth}
        \centering
        \includegraphics[width=\linewidth]{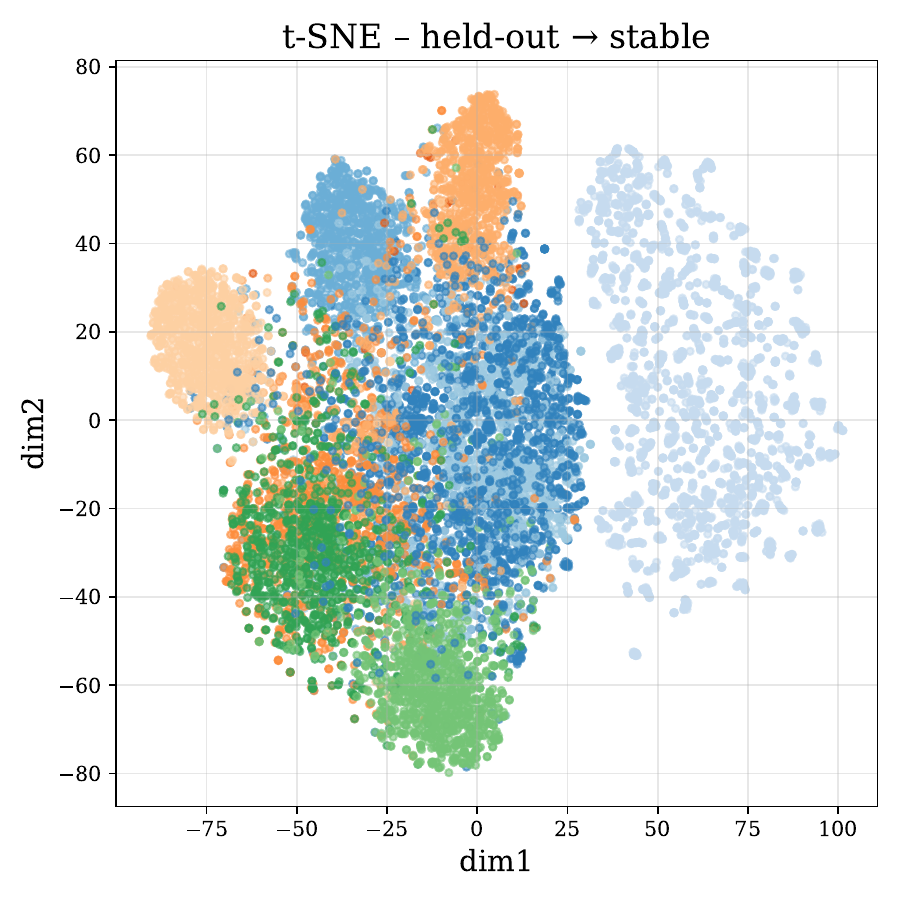}
        \subcaption{Held-out: Stable Diffusion}
    \end{minipage}
    \hfill
    \begin{minipage}[t]{0.32\textwidth}
        \centering
        \includegraphics[width=\linewidth]{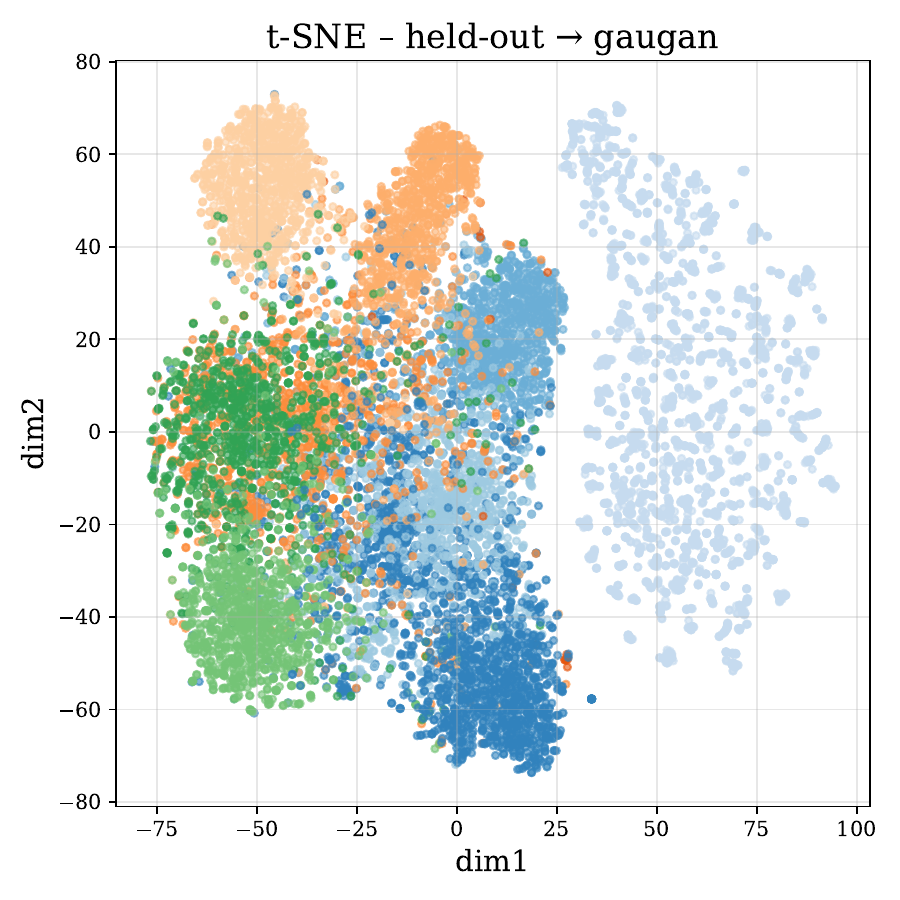}
        \subcaption{Held-out: GauGAN}
    \end{minipage}
    \hfill
    \begin{minipage}[t]{0.32\textwidth}
        \centering
        \includegraphics[width=\linewidth]{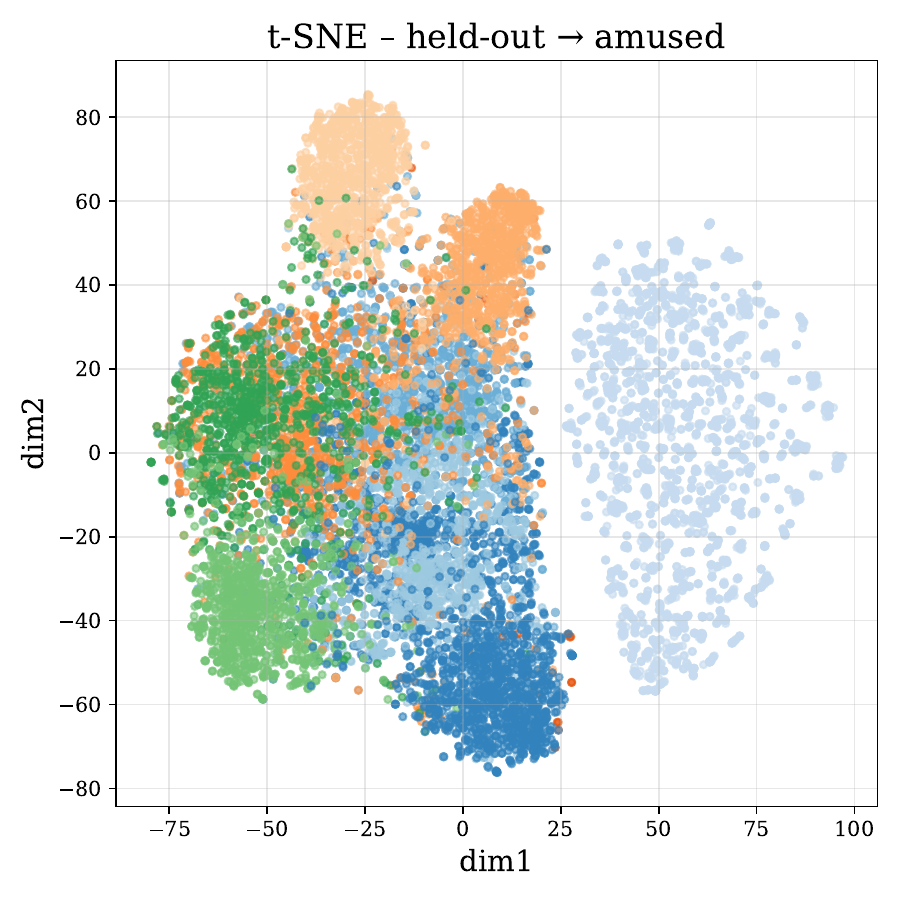}
        \subcaption{Held-out: Amused}
    \end{minipage}
    \caption{t-SNE projections of the embedding space after training. Light blue indicates the real samples. Real samples form compact clusters, while unseen generator samples (held out during training) remain distinct or peripheral.}
    \label{fig:tsne_trio}
\end{figure*}

\paragraph{Visualising Embedding Structure.}
To qualitatively assess the structure induced by our training procedure, we project the learned embeddings into 2D using t-SNE. Figure~\ref{fig:tsne_trio} shows the resulting layouts when individual generators are held out during training. Real samples form compact, well-separated clusters, while fake samples from unseen generators appear peripheral or distinct, illustrating UniGADD's ability to isolate anomalous inputs in feature space.

\section{Experiments}
\paragraph{Implementation details.}
We evaluate UniGADD on the AIGC datasets \citep{fan2024aigcbench}, covering diverse synthesis styles and content types. Images are resized to $224\times224$ and normalised with ImageNet statistics. Features are extracted using a frozen DINOv2 ViT-L/14 backbone \citep{oquab2024dinov2learningrobustvisual}, and a linear projection head maps embeddings to 512 dimensions. The projection is trained for 10 epochs with Adam ($\text{lr}=10^{-3}$), contrastive batch size 256, and temperature $\tau=0.07$. 

Real images are partitioned into clusters to provide pseudo-labels during training. We fix $n_{\text{total}}=40$ clusters, with $n_{\text{real}} = n_{\text{total}} - n_{\text{fake}}$ where $n_{\text{fake}}$ is the number of training generators. After training, class centroids are computed in the learned space and used for nearest-centroid classification at test time. The total number of clusters controls the granularity of the real-image manifold in the embedding space. A smaller number of clusters yields coarse partitions that may under-represent intra-class variation, while a larger number of clusters produces finer partitions that better capture local structure. In our setting, increasing the number of real clusters allows the model to more tightly characterise the nominal data distribution, which in turn improves separation at the boundary where unseen fake samples are likely to appear. Empirically, we observe that performance remains stable across a wide range of cluster counts (see Table \ref{tab:cluster_ablation}), indicating that the method is not overl   y sensitive to this hyperparameter. We therefore select a moderate value that balances representational fidelity and computational cost.

Evaluation follows a leave-one-out protocol: each generator is excluded in turn and used as an unseen test set, while the remaining generators and all real data form the training pool. Splits are sampled with a fixed random seed for reproducibility. All experiments are implemented in PyTorch and run on NVIDIA Tesla V100 GPUs.

\section{Results}
\label{sec:results}

\subsection{Evaluation Protocol}
\label{ssec:metrics}

\begin{figure*}[h!]
    \centering
     \includegraphics[width=1\columnwidth]{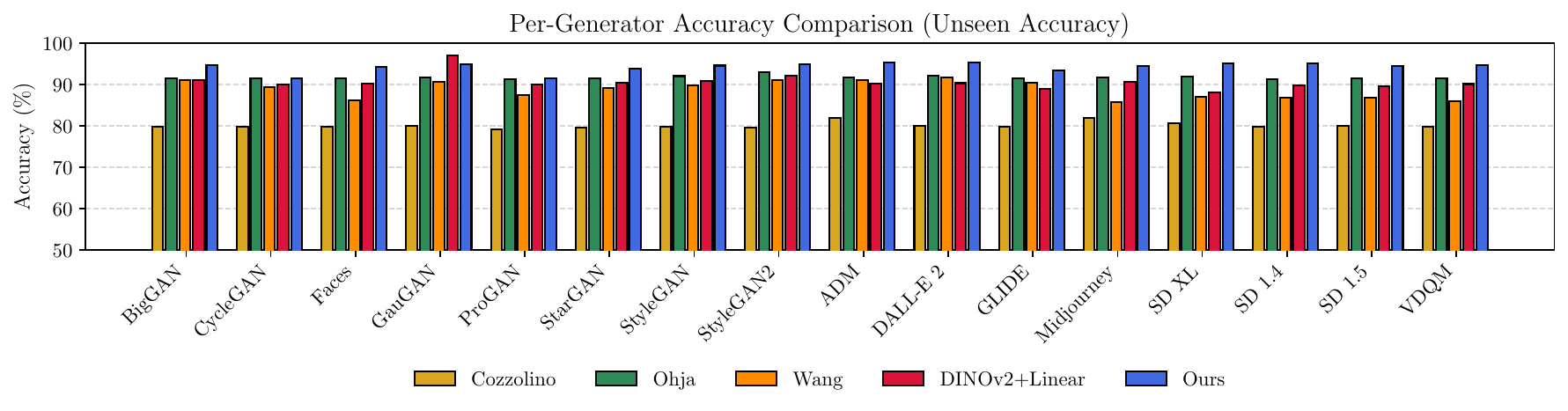}

    \includegraphics[width=1\columnwidth]{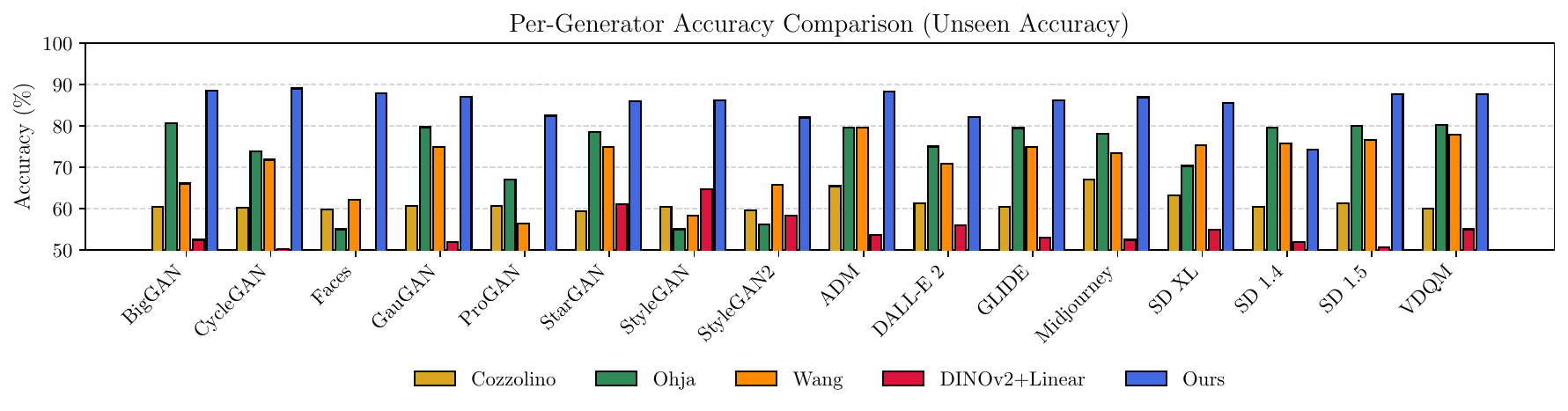}

    \caption{Average accuracy per generator under leave-one-out evaluation. Subfigure (a) shows accuracy on seen generators; (b) shows accuracy on unseen ones. UniGADD maintains high performance across both, outperforming previous methods in generalisation.}
    \label{fig:per_generator_accuracy}
\end{figure*}

We report results using class-averaged accuracy (\%) to ensure equal weighting across real and fake classes. To evaluate generalisation under distribution shift, we separate test data into \textit{seen} (generators included during training) and \textit{unseen} (generators excluded) subsets. This separation simulates realistic deployment conditions where previously unknown generation techniques may emerge.

We compare UniGADD against recent state-of-the-art deepfake detection methods, including supervised and unsupervised approaches~\citep{ojha2023towards,wang2023dire,cozzolino2024raisingbaraigeneratedimage}. We also benchmark a DINOv2 + Linear classifier baseline to demonstrate that our method improves upon standard frozen-feature detectors by incorporating contrastive structure and cluster refinement. Our experimental design progressively increases the challenge of generalisation. We begin with a Leave-One-Out (LOO) setup, withholding one generator at a time, then extend to a Leave-$N$-Out (LNO) evaluation where multiple generators are excluded simultaneously. Finally, we examine the effect of changing the backbone encoder to test whether UniGADD's robustness is architecture-dependent.

\begin{table*}[h!]
\scriptsize
\centering
\renewcommand{\arraystretch}{0.5} 
\setlength{\tabcolsep}{7.5pt} 

\begin{tabular}{llcccccccc>{\columncolor[HTML]{D9EAD3}}c>{\columncolor[HTML]{D9EAD3}}c}
\toprule
& \textbf{Method $\rightarrow$} 
& \multicolumn{2}{c}{\textbf{Cozzolino et al. 	$\uparrow$}} 
& \multicolumn{2}{c}{\textbf{Ohja et al. $\uparrow$}} 
& \multicolumn{2}{c}{\textbf{Wang et al. $\uparrow$}} 
& \multicolumn{2}{c}{\textbf{DINOv2 + Linear $\uparrow$}} 
& \multicolumn{2}{c}{\textbf{Ours $\uparrow$}} \\
\cmidrule(lr){3-4} \cmidrule(lr){5-6} \cmidrule(lr){7-8} \cmidrule(lr){9-10} \cmidrule(lr){11-12}

& \textbf{$\downarrow$ Generator} & \textit{Seen} & \textit{Unseen} & \textit{Seen} & \textit{Unseen} & \textit{Seen} & \textit{Unseen}
& \textit{Seen} & \textit{Unseen} & \textit{Seen} & \textit{Unseen} \\

\midrule

\multirow{8}{*}{\rotatebox{90}{\textbf{GANs}}} 
& BigGAN     & 79.92 & 60.47 & 91.49 & 80.79 & 91.03 & 66.12 & 91.16 & 52.50 & 94.80 & 88.50 \\
& CycleGAN   & 79.84  & 60.18 & 91.55 & 73.79 & 89.48 & 71.86 & 89.97 & 50.17 & 91.58  & 89.10 \\
& Faces     & 79.77 & 59.82 & 91.59 & 55.05 & 86.27 & 62.10 & 90.20 & 48.04 & 94.24 & 88.00 \\
& GauGAN     & 79.96 & 60.60 & 91.69 & 79.74 & 90.76 & 74.95 & 97.16 & 51.92 & 95.03 & 87.06 \\
& ProGAN     & 79.22 & 60.60 & 91.39 & 67.09 & 87.42 & 56.47 & 90.04 & 45.06 & 91.47 & 82.50 \\
& StarGAN    & 79.64 & 59.42 & 91.51 & 78.47 & 89.19 & 74.88 & 90.50 & 61.12 & 93.88 & 86.00 \\
& StyleGAN   & 79.93 & 60.53 & 92.09 & 55.05 & 89.94 & 58.41 & 90.89 & 64.73 & 94.64 & 86.20 \\
& StyleGAN2  & 79.67 & 59.53 & 93.13 & 56.21 & 91.06 & 65.83 & 92.10 & 58.29 & 94.87 & 82.08 \\

\midrule

\multirow{8}{*}{\rotatebox{90}{\textbf{Diffusion}}}
& ADM        & 81.98 & 65.47 & 91.67 & 79.68 & 91.06 & 79.63 & 90.32 & 53.64 & 95.30 & 88.32 \\
& DALL-E 2   & 80.14 & 61.28 & 92.23 & 75.05 & 91.77 & 70.85 & 90.37 & 56.06 & 95.30 & 82.21 \\
& GLIDE      & 79.93 & 60.48 & 91.55 & 79.53 & 90.58 & 74.90 & 89.03 & 53.12 & 93.42 & 86.23 \\
& Midjourney & 81.95 & 66.98 & 91.69 & 78.05 & 85.75 & 73.50 & 90.77 & 52.50 & 94.51 & 86.97 \\
& SD XL      & 80.78 & 63.27 & 91.88 & 70.37 & 87.10 & 75.28 & 88.10 & 55.00 & 95.13 & 85.51 \\
& SD 1.4     & 79.94 & 60.57 & 91.32 & 79.68 & 86.80 & 75.74 & 89.91 & 52.06 & 95.13 & 74.25 \\
& SD 1.5     & 80.15 & 61.28 & 91.55 & 80.05 & 86.96 & 76.64 & 89.58 & 50.75 & 94.61 & 87.75 \\
& VDQM       & 79.80 & 59.98 & 91.49 & 80.26 & 86.10 & 77.90 & 90.17 & 55.05 & 94.70 & 87.75 \\

\midrule

\rowcolor[HTML]{EFEFEF} \multirow{8}{*}{} & \textbf{AVG} & 80.16 & 59.82 & 91.68 & 73.64 & 88.62 & 70.79 & 90.32 & 53.84 & \cellcolor[HTML]{D9EAD3} 94.29 & \cellcolor[HTML]{D9EAD3} 85.53 \\

\rowcolor[HTML]{EFEFEF} \multirow{8}{*}{} & \textbf{Range $\downarrow$} & 2.76 & 22.50 & 0.91 & 25.74 & 5.68 & 21.43 & 9.06 & 19.67 & \cellcolor[HTML]{D9EAD3} 3.83 & \cellcolor[HTML]{D9EAD3} 14.85 \\

\bottomrule
\end{tabular}
\caption{Detection performance by generator across different methods for Leave-One-Out, showing \textit{Seen} and \textit{Unseen} accuracy. Generators are grouped by model type (GANs vs Diffusion). Our (UniGADD) results are highlighted in green.}
\label{tab:leave_one_out}
\end{table*}

\subsection{Leave-One-Out Generalisation}
\label{ssec:loo}

We first assess UniGADD's performance in the LOO setting. In each leave-one-out run, one generator is held out in turn, ensuring that every dataset serves once as the unseen test domain; within each split, train/test partitions are sampled randomly on multiple runs and averaged. Results are presented in Table~\ref{tab:leave_one_out} and visualised in Figure~\ref{fig:per_generator_accuracy}. It should be noted that the held-out sets include both GAN-based and diffusion-based models (e.g., excluding GauGAN while training on diffusion models, or vice versa), our protocol naturally covers cross-paradigm transfer between GANs and diffusion models.

Across all methods, performance on unseen generators is consistently lower than on seen ones, confirming the difficulty of out-of-distribution detection. However, UniGADD achieves the best unseen accuracy overall (\textbf{85.53\%}), significantly outperforming prior methods, including supervised (Ohja: 73.64\%) and unsupervised baselines (DINOv2+Linear: 53.84\%). UniGADD also exhibits the lowest variance across unseen generators (\textbf{range: 14.85}), suggesting consistent robustness even when exposed to unfamiliar generation styles. Notably, diffusion models—often harder to detect—see substantial improvements with UniGADD. For example, on Midjourney and ADM, our method achieves over \textbf{86\%} accuracy, whereas DINOv2+Linear drops below 55\%. These findings validate that the contrastive-prototype training and real-cluster refinement stages equip UniGADD with stronger resilience to distribution shift than previous approaches.

\begin{figure}[h!]
    \centering
    \includegraphics[width=0.6\columnwidth]{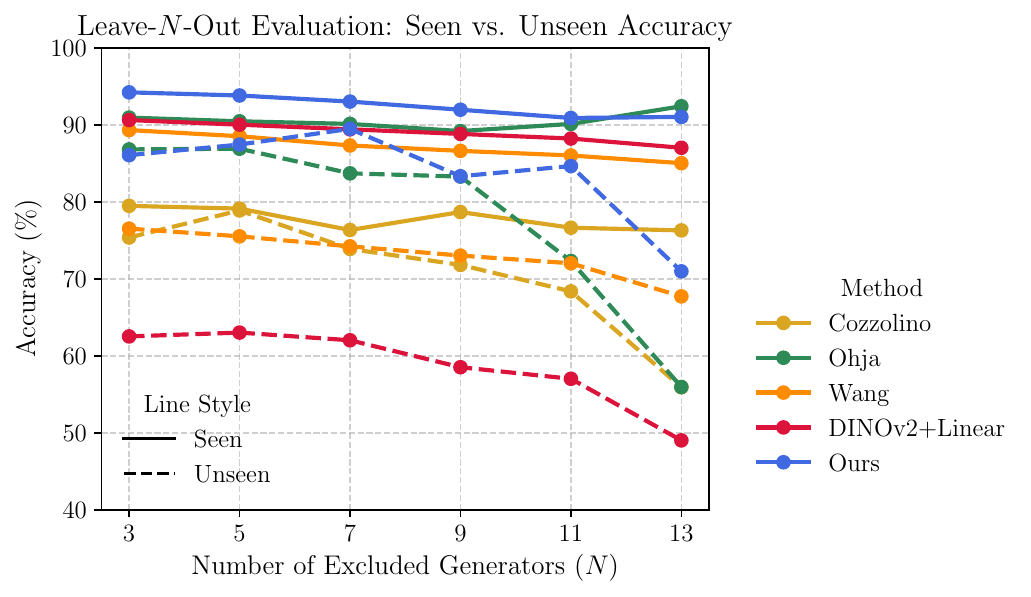}
    \caption{Average accuracy as the number of generators excluded from training increases. UniGADD generalises more robustly than prior methods.}
    \label{fig:NLeftOut}
\end{figure}
\subsection{Leave-$N$-Out Generalisation}
\label{ssec:lno}

To further evaluate generalisation under increasingly challenging scenarios, we conduct a Leave-$N$-Out evaluation, varying $N$ from 3 to 13. For each $N$, 10 random generator subsets are withheld from training and averaged to ensure stability. Results are summarised in Table~\ref{tab:leave_n_out} and visualised in Figure~\ref{fig:NLeftOut}.

As expected, all methods experience performance degradation as $N$ increases. However, UniGADD retains high accuracy on both seen (\textbf{92.47\%}) and unseen generators (\textbf{83.63\%} on average). Even with 13 generators excluded—nearly 80\% of all available training data—UniGADD outperforms others by large margins. Compared to Ohja et al., our method shows a smaller drop in unseen accuracy from $N=3$ to $N=13$ (\textbf{15.09pp} vs \textbf{30.86pp}), underscoring UniGADD's graceful degradation under extreme distributional shifts. The range for unseen accuracy is also lowest for UniGADD (\textbf{18.51}), again demonstrating stability across permutations. This experiment supports our hypothesis that refining the geometry of the real distribution during training leads to broader generalisation across diverse, previously unobserved generative mechanisms.

\begin{table*}[h!]
\scriptsize
\centering
\renewcommand{\arraystretch}{1} 
\setlength{\tabcolsep}{4pt} 

\begin{tabular}{clccccccc>{\columncolor[HTML]{D9EAD3}}c>{\columncolor[HTML]{D9EAD3}}c}
\toprule
 \textbf{Method $\rightarrow$} 
& \multicolumn{2}{c}{\textbf{Cozzolino et al. 	$\uparrow$}} 
& \multicolumn{2}{c}{\textbf{Ohja et al. $\uparrow$}} 
& \multicolumn{2}{c}{\textbf{Wang et al. $\uparrow$}} 
& \multicolumn{2}{c}{\textbf{DINOv2 + Linear $\uparrow$}} 
& \multicolumn{2}{c}{\textbf{Ours $\uparrow$}} \\

\cmidrule(lr){2-3} \cmidrule(lr){4-5} \cmidrule(lr){6-7} \cmidrule(lr){8-9} \cmidrule(lr){10-11}

 \textbf{$\downarrow$ $N$ Unseen} & \textit{Seen} & \textit{Unseen} & \textit{Seen} & \textit{Unseen} & \textit{Seen} & \textit{Unseen}
& \textit{Seen} & \textit{Unseen} & \textit{Seen} & \textit{Unseen} \\

\midrule

3   & 79.455 & 75.35 & 90.92 & 86.79 & 89.3 & 76.5 & 90.6 & 62.5 & 94.20 & 86.04 \\
5   & 79.10 & 78.86 & 90.44 & 86.88 & 88.5 & 75.5 & 90.0 & 63.0 & 93.79 & 87.42 \\
7   & 76.32 & 73.86 & 90.10 & 83.68 & 87.3 & 74.2 & 89.4 & 62.0 & 93.00 & 89.46 \\
9   & 78.65 & 71.79 & 89.17 & 83.26 & 86.6 & 73.0 & 88.8 & 58.5 & 91.94 & 83.30 \\
11  & 76.62 & 68.35 & 90.10 & 72.29 & 86.0 & 72.0 & 88.2 & 57.0 & 90.87 & 84.63 \\
13  & 76.27 & 55.90 & 92.39 & 55.93 & 85.0 & 67.7 & 87.0 & 49.0 & 91.01 & 70.95 \\

\midrule

\rowcolor[HTML]{EFEFEF} \multirow{8}{*}{} \textbf{AVG} & 77.74 & 73.73 & 90.52 & 78.14 & 87.12 & 73.15 & 89.00  & 58.67  & \cellcolor[HTML]{D9EAD3} 92.47 & \cellcolor[HTML]{D9EAD3} 83.63 \\

\rowcolor[HTML]{EFEFEF} \multirow{8}{*}{} \textbf{Range $\downarrow$} & 3.19 & 29.20 & 3.22 & 30.95 & 4.30 & 8.80  & 3.60 & 14.00 & \cellcolor[HTML]{D9EAD3} 3.33 & \cellcolor[HTML]{D9EAD3} 18.51 \\

\bottomrule

\end{tabular}
\caption{Detection performance by generator across different methods for Leave-$N$-Out, showing \textit{Seen} and \textit{Unseen} accuracy. Our (UniGADD) results are highlighted in green.}
\label{tab:leave_n_out}
\end{table*}


\subsection{Backbone Robustness Analysis}
\label{ssec:backbone}

To ensure our results are not backbone-specific, we perform ablations using four different pretrained vision encoders: CLIP ViT-B/32, CLIP ViT-L/14, DINOv2-S/14, and DINOv2-B/14 (our default). Table~\ref{tab:backbone_comparison} reports seen and unseen accuracy for each configuration under the LOO setting.

CLIP-based backbones, while strong on seen data (up to \textbf{96.72\%}), exhibit larger gaps on unseen samples, particularly for the smaller ViT-B/32 (only \textbf{70.57\%} unseen accuracy). In contrast, DINOv2-B/14 achieves better balance with \textbf{92.48\%} seen and \textbf{81.47\%} unseen accuracy, offering a compelling trade-off between closed- and open-set performance.

These results demonstrate that while UniGADD benefits from stronger backbone representations, its core generalisation ability holds across architectures. The refinement stage appears to amplify the strengths of foundation models while mitigating their overfitting tendencies. Further ablation studies analysing the sensitivity of UniGADD to clustering configuration, decision thresholds, backbone choice, and loss formulation are provided in Appendix \ref{app:ablations}.

\begin{table*}[h!]
\scriptsize
\centering
\renewcommand{\arraystretch}{0.5}
\setlength{\tabcolsep}{7.5pt}

\begin{tabular}{llcccccccc}
\toprule
& \textbf{Backbone $\rightarrow$} 
& \multicolumn{2}{c}{\textbf{ViT-B-32 $\uparrow$}} 
& \multicolumn{2}{c}{\textbf{ViT-L-14 $\uparrow$}} 
& \multicolumn{2}{c}{\textbf{DINOv2-B/14 $\uparrow$}} 
& \multicolumn{2}{c}{\textbf{DINOv2-S/14 $\uparrow$}} \\
\cmidrule(lr){3-4} \cmidrule(lr){5-6} \cmidrule(lr){7-8} \cmidrule(lr){9-10}

& \textbf{$\downarrow$ Generator} & \textit{Seen} & \textit{Unseen} & \textit{Seen} & \textit{Unseen}
& \textit{Seen} & \textit{Unseen} & \textit{Seen} & \textit{Unseen} \\

\midrule

\multirow{8}{*}{\rotatebox{90}{\textbf{GANs}}}
& BigGAN        & 89.54 & 62.00 & 96.55 & 88.50 & 91.02 & 91.25 & 81.91 & 65.75 \\
& CycleGAN      & 87.57 & 71.50 & 96.68 & 88.50 & 92.83 & 84.25 & 79.93 & 69.50 \\
& Faces         & 86.74 & 57.25 & 96.68 & 86.50 & 89.70 & 76.25 & 79.11 & 63.25 \\
& GauGAN        & 90.72 & 67.75 & 96.38 & 83.50 & 91.41 & 87.25 & 82.47 & 68.25 \\
& ProGAN        & 89.12 & 76.25 & 96.91 & 81.25 & 94.22 & 77.92 & 81.56 & 76.25 \\
& StarGAN       & 87.96 & 66.25 & 96.88 & 88.25 & 90.20 & 89.50 & 80.10 & 78.50 \\
& StyleGAN      & 89.80 & 62.75 & 97.01 & 81.25 & 94.24 & 70.00 & 82.30 & 58.00 \\
& StyleGAN2     & 87.07 & 63.75 & 96.32 & 87.75 & 94.54 & 81.75 & 78.98 & 61.00 \\

\midrule

\multirow{8}{*}{\rotatebox{90}{\textbf{Diffusion}}}
& ADM           & 89.80 & 70.00 & 96.97 & 87.75 & 93.12 & 87.75 & 81.25 & 66.00 \\
& DALL-E 2      & 89.21 & 70.50 & 96.58 & 87.50 & 93.06 & 77.75 & 86.25 & 74.25 \\
& GLIDE         & 87.60 & 76.25 & 96.74 & 88.25 & 89.51 & 74.25 & 83.29 & 59.50 \\
& Midjourney    & 92.57 & 78.00 & 97.24 & 87.25 & 93.88 & 81.50 & 85.10 & 76.50 \\
& SD XL         & 92.73 & 80.25 & 96.84 & 87.00 & 91.68 & 79.25 & 81.55 & 72.50 \\
& SD 1.4        & 88.82 & 80.25 & 96.78 & 87.00 & 94.38 & 86.25 & 83.85 & 79.25 \\
& SD 1.5        & 90.26 & 81.25 & 96.64 & 84.75 & 92.57 & 83.25 & 82.17 & 74.50 \\
& VDQM          & 89.84 & 69.25 & 96.81 & 85.50 & 92.96 & 86.50 & 82.83 & 67.00 \\

\midrule

\rowcolor[HTML]{EFEFEF}
& \textbf{AVG} & 89.15 & 70.57 & 96.72 & 86.53 & 92.48 & 81.47 & 81.86 & 68.56 \\

\rowcolor[HTML]{EFEFEF}
& \textbf{Range $\downarrow$} & 6.00 & 24.00 & 1.69 & 7.25 & 5.03 & 21.25 & 7.29 & 21.50 \\

\bottomrule
\end{tabular}
\caption{Detection accuracy by backbone across GAN and Diffusion generators under Leave-One-Out setting.}
\label{tab:backbone_comparison}
\end{table*}

\subsection{Limitations}

UniGADD shows strong generalisation to unseen generators, but it has several limitations. The method relies on frozen vision transformer backbones such as DINOv2 and CLIP, so any biases or weaknesses in these models are inherited. For example, detection may be less reliable under unusual lighting conditions or non-photographic image styles. The current method only uses single images and does not capture temporal signals that can be useful for spotting subtle video forgeries. Extending UniGADD to video is an important direction for future work.

Our experiments are based on balanced datasets with a reasonable number of samples per generator. The performance of UniGADD in low-shot settings, highly imbalanced data, or in-the-wild content remains untested. Although inference is fast and lightweight, the clustering step used for prototype refinement becomes more expensive as the dataset grows. Large-scale deployments may require streaming or incremental clustering methods.

We do not explicitly evaluate robustness to low-level image perturbations such as compression or blur, which remain important directions for extending our approach.

Finally, we use a fixed threshold for classification, selected on a held-out validation set. In practical deployments, the decision boundary may need to adapt over time as content distributions shift. These points highlight areas where further refinement is needed before UniGADD is suitable for production-level deployment.

\section{Discussion \& Conclusion}

We introduced UniGADD, a generator-agnostic deepfake detector that performs well on both seen and unseen manipulations. The method uses a two-stage approach. First, supervised contrastive learning builds a structured embedding space using real and labelled fake examples. Then, a refinement step improves the compactness of real-image clusters, making it easier to detect outliers caused by new manipulations. Results show that UniGADD matches the accuracy of other detectors on known generators and outperforms them when generalising to novel ones. It maintains high accuracy even as more generators are excluded from training. The method also shows low variance across different settings, which suggests consistent performance.

We tested UniGADD with different backbone models and found that the benefits of refinement hold across architectures. This supports the idea that UniGADD improves generalisation without relying on any one encoder. Because it operates in a frozen feature space, it also avoids the cost of full fine-tuning. Future work will explore adding temporal features for video, testing the method on fewer training samples, and using adversarial training to increase robustness. These changes could make UniGADD more useful in open-world conditions.

UniGADD offers a simple, effective, and interpretable approach to deepfake detection. Its prototype-based decisions are easy to understand and adapt as new content appears.

\bibliography{main}

@String(ICCVW = {IEEE/CVF International Conference on Computer Vision Workshops})

@String(ICML = {International Conference on Machine Learning})

@String(CVPR= {IEEE Conf. Comput. Vis. Pattern Recog.})

@String(ICCV= {Int. Conf. Comput. Vis.})

@String(ECCV= {Eur. Conf. Comput. Vis.})

@String(ICASSP=	{ICASSP})

@String(IJCAI = {IJCAI})

@String(CVPRW= {IEEE Conf. Comput. Vis. Pattern Recog. Worksh.})

@String(CVPR  = {CVPR})

@String(ICCV  = {ICCV})

@String(ECCV  = {ECCV})

@String(CVPRW= {CVPRW})

@Article{DeepfakeAttacks,
AUTHOR = {Naitali, Amal and Ridouani, Mohammed and Salahdine, Fatima and Kaabouch, Naima},
TITLE = {Deepfake Attacks: Generation, Detection, Datasets, Challenges, and Research Directions},
JOURNAL = {Computers},
VOLUME = {12},
YEAR = {2023},
NUMBER = {10},
ARTICLE-NUMBER = {216},
URL = {https://www.mdpi.com/2073-431X/12/10/216},
ISSN = {2073-431X},
}

@article{DeepfakeSurvey,
journal = {ACM Computing Surveys},
author = {Wang, Tianyi and Liao, Xin and Chow, Kam Pui and Lin, Xiaodong and Wang, Yinglong},
title = {Deepfake Detection: A Comprehensive Survey from the Reliability Perspective},
year = {2024},
issue_date = {March 2025},
publisher = {Association for Computing Machinery},
address = {New York, NY, USA},
volume = {57},
number = {3},
issn = {0360-0300},
url = {https://doi.org/10.1145/3699710},
doi = {10.1145/3699710},
}

@misc{faceforensics,
      title={FaceForensics++: Learning to Detect Manipulated Facial Images}, 
      author={Andreas Rössler and Davide Cozzolino and Luisa Verdoliva and Christian Riess and Justus Thies and Matthias Nießner},
      year={2019},
      eprint={1901.08971},
      archivePrefix={arXiv},
      primaryClass={cs.CV},
      url={https://arxiv.org/abs/1901.08971}, 
}

@article{zhu2023genimage,
  title={Genimage: A million-scale benchmark for detecting ai-generated image},
  author={Zhu, Mingjian and Chen, Hanting and Yan, Qiangyu and Huang, Xudong and Lin, Guanyu and Li, Wei and Tu, Zhijun and Hu, Hailin and Hu, Jie and Wang, Yunhe},
  journal={Advances in Neural Information Processing Systems},
  volume={36},
  pages={77771--77782},
  year={2023}
}

@article{fan2024aigcbench,
  title={AIGCBench: Comprehensive evaluation of image-to-video content generated by AI},
  author={Fan, Fanda and Luo, Chunjie and Gao, Wanling and Zhan, Jianfeng},
  journal={BenchCouncil Transactions on Benchmarks, Standards and Evaluations},
  pages={100152},
  year={2024},
  publisher={Elsevier}
}

@misc{roth2022totalrecallindustrialanomaly,
      title={Towards Total Recall in Industrial Anomaly Detection}, 
      author={Karsten Roth and Latha Pemula and Joaquin Zepeda and Bernhard Schölkopf and Thomas Brox and Peter Gehler},
      year={2022},
      eprint={2106.08265},
      archivePrefix={arXiv},
      primaryClass={cs.CV},
      url={https://arxiv.org/abs/2106.08265}, 
}

@misc{defard2020padimpatchdistributionmodeling,
      title={PaDiM: a Patch Distribution Modeling Framework for Anomaly Detection and Localization}, 
      author={Thomas Defard and Aleksandr Setkov and Angelique Loesch and Romaric Audigier},
      year={2020},
      eprint={2011.08785},
      archivePrefix={arXiv},
      primaryClass={cs.CV},
      url={https://arxiv.org/abs/2011.08785}, 
}

@misc{oquab2024dinov2learningrobustvisual,
      title={DINOv2: Learning Robust Visual Features without Supervision}, 
      author={Maxime Oquab and Timothée Darcet and Théo Moutakanni and Huy Vo and Marc Szafraniec and Vasil Khalidov and Pierre Fernandez and Daniel Haziza and Francisco Massa and Alaaeldin El-Nouby and Mahmoud Assran and Nicolas Ballas and Wojciech Galuba and Russell Howes and Po-Yao Huang and Shang-Wen Li and Ishan Misra and Michael Rabbat and Vasu Sharma and Gabriel Synnaeve and Hu Xu and Hervé Jegou and Julien Mairal and Patrick Labatut and Armand Joulin and Piotr Bojanowski},
      year={2024},
      eprint={2304.07193},
      archivePrefix={arXiv},
      primaryClass={cs.CV},
      url={https://arxiv.org/abs/2304.07193}, 
}

@misc{sun2023contrastivepseudolearningopenworld,
      title={Contrastive Pseudo Learning for Open-World DeepFake Attribution}, 
      author={Zhimin Sun and Shen Chen and Taiping Yao and Bangjie Yin and Ran Yi and Shouhong Ding and Lizhuang Ma},
      year={2023},
      eprint={2309.11132},
      archivePrefix={arXiv},
      primaryClass={cs.CV},
      url={https://arxiv.org/abs/2309.11132}, 
}

@misc{larue2023seeablesoftdiscrepanciesbounded,
      title={SeeABLE: Soft Discrepancies and Bounded Contrastive Learning for Exposing Deepfakes}, 
      author={Nicolas Larue and Ngoc-Son Vu and Vitomir Struc and Peter Peer and Vassilis Christophides},
      year={2023},
      eprint={2211.11296},
      archivePrefix={arXiv},
      primaryClass={cs.CV},
      url={https://arxiv.org/abs/2211.11296}, 
}

@misc{han2024generalvideobaseddeepfakedetection,
      title={Towards More General Video-based Deepfake Detection through Facial Feature Guided Adaptation for Foundation Model}, 
      author={Yue-Hua Han and Tai-Ming Huang and Shu-Tzu Lo and Po-Han Huang and Kai-Lung Hua and Jun-Cheng Chen},
      year={2024},
      eprint={2404.05583},
      archivePrefix={arXiv},
      primaryClass={cs.CV},
      url={https://arxiv.org/abs/2404.05583}, 
}

@misc{shi2023realfacefoundationrepresentation,
      title={Real Face Foundation Representation Learning for Generalized Deepfake Detection}, 
      author={Liang Shi and Jie Zhang and Shiguang Shan},
      year={2023},
      eprint={2303.08439},
      archivePrefix={arXiv},
      primaryClass={cs.CV},
      url={https://arxiv.org/abs/2303.08439}, 
}

@ARTICLE{deepfakeSurvey2,
  author={Rana, Md Shohel and Nobi, Mohammad Nur and Murali, Beddhu and Sung, Andrew H.},
  journal={IEEE Access}, 
  title={Deepfake Detection: A Systematic Literature Review}, 
  year={2022},
  volume={10},
  number={},
  pages={25494-25513},
  doi={10.1109/ACCESS.2022.3154404}}

@InProceedings{Pellicer_2024_CVPR,
    author    = {Pellicer, Alvaro Lopez and Li, Yi and Angelov, Plamen},
    title     = {PUDD: Towards Robust Multi-modal Prototype-based Deepfake Detection},
    booktitle = {Proceedings of the IEEE/CVF Conference on Computer Vision and Pattern Recognition (CVPR) Workshops},
    month     = {June},
    year      = {2024},
    pages     = {3809-3817}
}

@misc{bird2023cifakeimageclassificationexplainable,
      title={CIFAKE: Image Classification and Explainable Identification of AI-Generated Synthetic Images}, 
      author={Jordan J. Bird and Ahmad Lotfi},
      year={2023},
      eprint={2303.14126},
      archivePrefix={arXiv},
      primaryClass={cs.CV},
      url={https://arxiv.org/abs/2303.14126}, 
}

@misc{radford2021learningtransferablevisualmodels,
      title={Learning Transferable Visual Models From Natural Language Supervision}, 
      author={Alec Radford and Jong Wook Kim and Chris Hallacy and Aditya Ramesh and Gabriel Goh and Sandhini Agarwal and Girish Sastry and Amanda Askell and Pamela Mishkin and Jack Clark and Gretchen Krueger and Ilya Sutskever},
      year={2021},
      eprint={2103.00020},
      archivePrefix={arXiv},
      primaryClass={cs.CV},
      url={https://arxiv.org/abs/2103.00020}, 
}

@misc{clip,
      title={Learning Transferable Visual Models From Natural Language Supervision}, 
      author={Alec Radford and Jong Wook Kim and Chris Hallacy and Aditya Ramesh and Gabriel Goh and Sandhini Agarwal and Girish Sastry and Amanda Askell and Pamela Mishkin and Jack Clark and Gretchen Krueger and Ilya Sutskever},
      year={2021},
      eprint={2103.00020},
      archivePrefix={arXiv},
      primaryClass={cs.CV},
      url={https://arxiv.org/abs/2103.00020}, 
}

@book{farid2016photo,
  title={Photo forensics},
  author={Farid, Hany},
  year={2016},
  publisher={MIT press}
}

@book{sencar2013digital,
  title={Digital image forensics},
  author={Sencar, Husrev T and Memon, Nasir},
  year={2013},
  publisher={Springer}
}

@inproceedings{yang2019exposing,
  title={{Exposing Deep Fakes Using Inconsistent Head Poses}},
  author={Yang, Xin and Li, Yuezun and Lyu, Siwei},
  booktitle=ICASSP,
  year={2019},
}

@inproceedings{ijcai2020p476,
  title     = {{FakeSpotter: A Simple yet Robust Baseline for Spotting AI-Synthesized Fake Faces}},
  author    = {Wang, Run and Juefei-Xu, Felix and Ma, Lei and Xie, Xiaofei and Huang, Yihao and Wang, Jian and Liu, Yang},
  booktitle = IJCAI,
  year      = {2020},
}

@inproceedings{sha2022fake,
  title={{DE-FAKE: Detection and Attribution of Fake Images Generated by Text-to-Image Generation Models}},
  author={Sha, Zeyang and Li, Zheng and Yu, Ning and Zhang, Yang},
  booktitle={ACM CCS},
  year={2023}
}

@inproceedings{ojha2023towards,
  title={{Towards Universal Fake Image Detectors That Generalize Across Generative Models}},
  author={Ojha, Utkarsh and Li, Yuheng and Lee, Yong Jae},
  booktitle=CVPR,
  year={2023}
}

@inproceedings{epstein2023online,
  title={{Online Detection of AI-Generated Images}},
  author={Epstein, David C and Jain, Ishan and Wang, Oliver and Zhang, Richard},
  booktitle=ICCVW,
  year={2023}
}

@InProceedings{wang2023dire,
  title={{DIRE for Diffusion-Generated Image Detection}},
  author={Wang, Zhendong and Bao, Jianmin and Zhou, Wengang and Wang, Weilun and Hu, Hezhen and Chen, Hong and Li, Houqiang},
  booktitle=ICCV,
  year={2023}
}

@inproceedings{baraldi2024contrasting,
  title={Contrasting deepfakes diffusion via contrastive learning and global-local similarities},
  author={Baraldi, Lorenzo and Cocchi, Federico and Cornia, Marcella and Baraldi, Lorenzo and Nicolosi, Alessandro and Cucchiara, Rita},
  booktitle=ECCV,
  year={2024},
}

@inproceedings{wang2020cnn,
  title={{CNN-Generated Images Are Surprisingly Easy to Spot... for Now}},
  author={Wang, Sheng-Yu and Wang, Oliver and Zhang, Richard and Owens, Andrew and Efros, Alexei A},
  booktitle=CVPR,
  year={2020}
}

@inproceedings{frank2020leveraging,
  title={{Leveraging frequency analysis for deep fake image recognition}},
  author={Frank, Joel and Eisenhofer, Thorsten and Sch{\"o}nherr, Lea and Fischer, Asja and Kolossa, Dorothea and Holz, Thorsten},
  booktitle=ICML,
  year={2020}
}

@InProceedings{Corvi_2023_CVPR,
    author    = {Corvi, Riccardo and Cozzolino, Davide and Poggi, Giovanni and Nagano, Koki and Verdoliva, Luisa},
    title     = {{Intriguing Properties of Synthetic Images: From Generative Adversarial Networks to Diffusion Models}},
    booktitle = CVPRW,
    year      = {2023},
}

@InProceedings{corvi2023detection,
  title={{On the detection of synthetic images generated by diffusion models}},
  author={Corvi, Riccardo and Cozzolino, Davide and Zingarini, Giada and Poggi, Giovanni and Nagano, Koki and Verdoliva, Luisa},
  booktitle=ICASSP,
  year={2023},
}

@article{khosla2020supervisedcontrastivelearning,
  title={Supervised Contrastive Learning},
  author={Prannay Khosla and Piotr Teterwak and Chen Wang and Aaron Sarna and Yonglong Tian and Phillip Isola and Aaron Maschinot and Ce Liu and Dilip Krishnan},
  journal={ArXiv},
  year={2020},
  volume={abs/2004.11362},
  url={https://api.semanticscholar.org/CorpusID:216080787}
}

@inproceedings{Snell2017PrototypicalNF,
  title={Prototypical Networks for Few-shot Learning},
  author={Jake Snell and Kevin Swersky and Richard S. Zemel},
  booktitle={Neural Information Processing Systems},
  year={2017},
  url={https://api.semanticscholar.org/CorpusID:309759}
}

@inproceedings{Chen2018ThisLL,
  title={This looks like that: deep learning for interpretable image recognition},
  author={Chaofan Chen and Oscar Li and Alina Jade Barnett and Jonathan Su and Cynthia Rudin},
  booktitle={Neural Information Processing Systems},
  year={2018},
  url={https://api.semanticscholar.org/CorpusID:49482223}
}

@article{ANGELOV2025129464,
title = {IDEAL: Interpretable-by-Design ALgorithms for learning from foundation feature spaces},
journal = {Neurocomputing},
volume = {626},
pages = {129464},
year = {2025},
issn = {0925-2312},
doi = {https://doi.org/10.1016/j.neucom.2025.129464},
url = {https://www.sciencedirect.com/science/article/pii/S0925231225001365},
author = {Plamen Angelov and Dmitry Kangin and Ziyang Zhang}
}

@article{Donnelly2021DeformablePA,
  title={Deformable ProtoPNet: An Interpretable Image Classifier Using Deformable Prototypes},
  author={Jonathan Donnelly and Alina Jade Barnett and Chaofan Chen},
  journal={2022 IEEE/CVF Conference on Computer Vision and Pattern Recognition (CVPR)},
  year={2021},
  pages={10255-10265},
  url={https://api.semanticscholar.org/CorpusID:244729769}
}

@article{Wu2021ProtoTransformerAM,
  title={ProtoTransformer: A Meta-Learning Approach to Providing Student Feedback},
  author={Mike Wu and Noah D. Goodman and Chris Piech and Chelsea Finn},
  journal={ArXiv},
  year={2021},
  volume={abs/2107.14035},
  url={https://api.semanticscholar.org/CorpusID:236493494}
}

@article{Caron2021EmergingPI,
  title={Emerging Properties in Self-Supervised Vision Transformers},
  author={Mathilde Caron and Hugo Touvron and Ishan Misra and Herv'e J'egou and Julien Mairal and Piotr Bojanowski and Armand Joulin},
  journal={2021 IEEE/CVF International Conference on Computer Vision (ICCV)},
  year={2021},
  pages={9630-9640},
  url={https://api.semanticscholar.org/CorpusID:233444273}
}

@article{Caron2020UnsupervisedLO,
  title={Unsupervised Learning of Visual Features by Contrasting Cluster Assignments},
  author={Mathilde Caron and Ishan Misra and Julien Mairal and Priya Goyal and Piotr Bojanowski and Armand Joulin},
  journal={ArXiv},
  year={2020},
  volume={abs/2006.09882},
  url={https://api.semanticscholar.org/CorpusID:219721240}
}

@article{Weng2025ClusteringPO,
  title={Clustering Properties of Self-Supervised Learning},
  author={Xi Weng and Jianing An and Xudong Ma and Binhang Qi and Jie Luo and Xi Yang and Jin Song Dong and Lei Huang},
  journal={ArXiv},
  year={2025},
  volume={abs/2501.18452},
  url={https://api.semanticscholar.org/CorpusID:275993509}
}

@misc{cozzolino2024raisingbaraigeneratedimage,
      title={Raising the Bar of AI-generated Image Detection with CLIP}, 
      author={Davide Cozzolino and Giovanni Poggi and Riccardo Corvi and Matthias Nießner and Luisa Verdoliva},
      year={2024},
      eprint={2312.00195},
      archivePrefix={arXiv},
      primaryClass={cs.CV},
      url={https://arxiv.org/abs/2312.00195}, 
}
\bibliographystyle{tmlr}

\newpage

\appendix
\section{Additional Ablation Studies}
\label{app:ablations}

\begin{table}[h!]
\small
\centering
\renewcommand{\arraystretch}{1}
\setlength{\tabcolsep}{6pt}

\begin{tabular}{cccccc}
\toprule
\textbf{Clusters} & \textbf{$n_{\text{real}}$} & \textbf{Ratio} & \textbf{Seen} & \textbf{Unseen} \\
\midrule
20  & 3  & 0.19 & 87.6 & 45.8 \\
30  & 13 & 0.81 & 87.0 & 54.8 \\
40  & 23 & 1.44 & 87.4 & 54.0 \\
60  & 43 & 2.69 & 88.4 & 54.5 \\
80  & 63 & 3.94 & 88.3 & 52.2 \\
\rowcolor[HTML]{EFEFEF}
100 & 83 & 5.19 & \textbf{88.6} & \textbf{56.6} \\
\bottomrule
\end{tabular}

\caption{Cluster ablation showing the effect of varying the total number of clusters.}
\label{tab:cluster_ablation}
\end{table}


\begin{table}[h!]
\small
\centering
\renewcommand{\arraystretch}{1}
\setlength{\tabcolsep}{6pt}

\begin{tabular}{cccc}
\toprule
\textbf{$\sigma$} & \textbf{Seen Acc} & \textbf{Unseen Acc} \\
\midrule
1.0 & 91.5 & 46.2 \\
1.5 & 90.4 & 51.3 \\
2.0 & 89.7 & 54.0 \\
2.5 & 89.3 & 55.2 \\
3.0 & 89.2 & 55.7 \\
3.5 & 89.1 & 56.0 \\
\rowcolor[HTML]{EFEFEF}
4.0 & 89.0 & \textbf{56.2} \\
\bottomrule
\end{tabular}

\caption{Threshold ablation showing the effect of varying $\sigma$ on detection accuracy.}
\label{tab:threshold_ablation}
\end{table}

\begin{table}[h!]
\small
\centering
\renewcommand{\arraystretch}{1}
\setlength{\tabcolsep}{6pt}

\begin{tabular}{ccc}
\toprule
\textbf{Backbone} & \textbf{Weighted Acc} & \textbf{Silhouette} \\
\midrule
ViT-B-32      & 80.10 & \textbf{0.328} \\
ViT-L-14      & \textbf{83.42} & 0.312 \\
DINOv2-B/14   & 82.66 & 0.293 \\
DINOv2-S/14   & 81.64 & 0.278 \\
\bottomrule
\end{tabular}

\caption{Extended backbone comparison showing weighted accuracy and embedding quality (silhouette score).}
\label{tab:backbone_extended}
\end{table}

\begin{table}[h!]
\small
\centering
\renewcommand{\arraystretch}{1}
\setlength{\tabcolsep}{6pt}

\begin{tabular}{ccc}
\toprule
\textbf{Loss Configuration} & \textbf{Seen} & \textbf{Unseen} \\
\midrule
$L_{\text{contrast}} + L_{\text{sp}}$     & 89.1 & 71.4 \\
\rowcolor[HTML]{EFEFEF}
$L_{\text{contrast}} + L_{\text{wp}}$     & \textbf{91.4} & \textbf{73.0} \\
$L_{\text{contrast}} + L_{\text{refine}}$ & 87.9 & 61.2 \\
\bottomrule
\end{tabular}

\caption{Loss ablation comparing different refinement strategies.}
\label{tab:loss_ablation}
\end{table}

\end{document}